\documentclass[10pt, a4paper]{article}
\usepackage[scientific-notation=true]{siunitx}
\usepackage{subcaption}
\usepackage{array}
\usepackage{url}
\usepackage[dvipsnames]{xcolor}
\usepackage{colortbl}
\usepackage{tikz}
\usepackage{booktabs}
\usepackage{cellspace}
\usepackage[margin=1in]{geometry}
\usepackage{caption}
\usepackage{multicol}
\usepackage{float}
\usepackage[export]{adjustbox} 
\usepackage[inkscapelatex=false]{svg} 
\usepackage{pgfplots}
\usepackage[most]{tcolorbox}
\usepackage[dvipsnames]{xcolor}
\usepackage{enumitem}
\usepackage{tabularx}
\usepackage[toc,page]{appendix}
\usepackage{multirow}
\usepackage{todonotes}

\tcbuselibrary{skins,breakable,raster,most}
\definecolor{latxaC}{RGB}{ 90,183,183}
\definecolor{llamaC}{RGB}{121,137,255}
\definecolor{qwenC} {RGB}{172,172,172}
\definecolor{earlyC}{RGB}{215,150,200}
\definecolor{customblue}{RGB}{70,130,180}  

\usepackage[final]{lrec2026} 

\title{Multimodal Large Language Models for Low-Resource Languages: A Case Study for Basque}

\name{Lukas Arana, Julen Etxaniz, Ander Salaberria, Gorka Azkune} 

\address{HiTZ Basque Center for Language Technology - Ixa NLP Group, University of the Basque Country UPV/EHU\\
         \{lukas.arana, julen.etxaniz, ander.salaberria, gorka.azkune\}@ehu.eus\\}

\abstract{
Current Multimodal Large Language Models exhibit very strong performance for several demanding tasks. While commercial MLLMs deliver acceptable performance in low-resource languages, comparable results remain unattained within the open science community. In this paper, we aim to develop a strong MLLM for a low-resource language, namely Basque. For that purpose, we develop our own training and evaluation image-text datasets. Using two different Large Language Models as backbones, the Llama-3.1-Instruct model and a Basque-adapted variant called Latxa, we explore several data mixtures for training. We show that: i) low ratios of Basque multimodal data (around 20\%) are already enough to obtain solid results on Basque benchmarks, and ii) contrary to expected, a Basque instructed backbone LLM is not required to obtain a strong MLLM in Basque.  Our results pave the way to develop MLLMs for other low-resource languages by openly releasing our resources. 
 \\ \newline  }
\begin{document}

\maketitleabstract

\section{Introduction}
\label{sec:introduction}

Multimodal Large Language Models (MLLMs) \cite{MLLMReview} aim to further improve assistance systems by combining textual data with other information modalities, such as images, video, or audio. In the context of MLLMs designed for text and image processing, these systems can perform novel tasks that traditional text-only Large Language Models (LLMs) cannot natively support, such as image captioning, visual question answering, or optical character recognition, among others. Due to these capabilities, the majority of the most advanced LLMs are being shared as natively multimodal \cite{gpt5, geminipro}.


However, current MLLMs are primarily trained with English resources and thus still face performance degradation in low-resource languages. This aspect has been thoroughly studied \cite{pangea}, showing the multilingual performance gap between modern proprietary models and open-weight alternatives. Despite recent efforts to improve the multilingual capabilities of MLLMs \cite{ayavision}, there is still a significant performance gap between open-weight and proprietary systems, especially in the context of low-resource languages. 

Developing an MLLM requires addressing a vast number of design decisions \cite{MLLMReview}, many of which have not been properly explored for low-resource languages. This paper proposes to study the development and performance of various training recipes and evaluation methods to build \textbf{the first open MLLM for Basque}. Although centred on a single language, our study likely generalizes to other similarly resourced languages. Basque is a low-resource language that lacks multimodal datasets and ranks around 50th in Common Crawl with roughly 1,000× less text data than English. 


In this paper, we create \textbf{the first multimodal datasets for Basque}, both for training and evaluation. We only use open resources, avoiding the use of proprietary systems. We rely on translation procedures specifically adapted to the requirements of the datasets. As a result, we generate more than 3 million image-text instances for training and around 8 thousand evaluation instances from human-validated benchmarks\footnote{https://huggingface.co/collections/HiTZ/multimodal-latxa}.

With our new datasets, we run a systematic exploration with MLLMs, following the late-fusion paradigm \cite{nvlm} to adapt two pretrained LLMs for multimodal tasks: English-centric Llama-3.1-8B-Instruct \cite{llama} and Basque instructed Latxa-Llama-3.1-8B-Instruct \cite{latxa}. In our experiments, we find that:


\textbf{1) Low ratios of Basque multimodal data are enough to build a strong MLLM for Basque.} Using only 20\% of the training data in Basque (the rest in English), we already achieve a very performant MLLM for our Basque benchmarks. Furthermore, our experiments suggest that English multimodal data could be enough for a decent performance, assuming that text-only Basque data can be used for training. The implications of this finding are important, since obtaining aligned image-text data for a low-resource language is generally more difficult than text-only data.

\textbf{2) A Basque instructed backbone LLM is not required to build a strong MLLM for Basque.} Contrary to expectations, an English-centric backbone LLM can achieve the same performance, even for open-ended text generation.

\begin{figure*}[ht]
    \centering
    \includegraphics[width=1\linewidth]{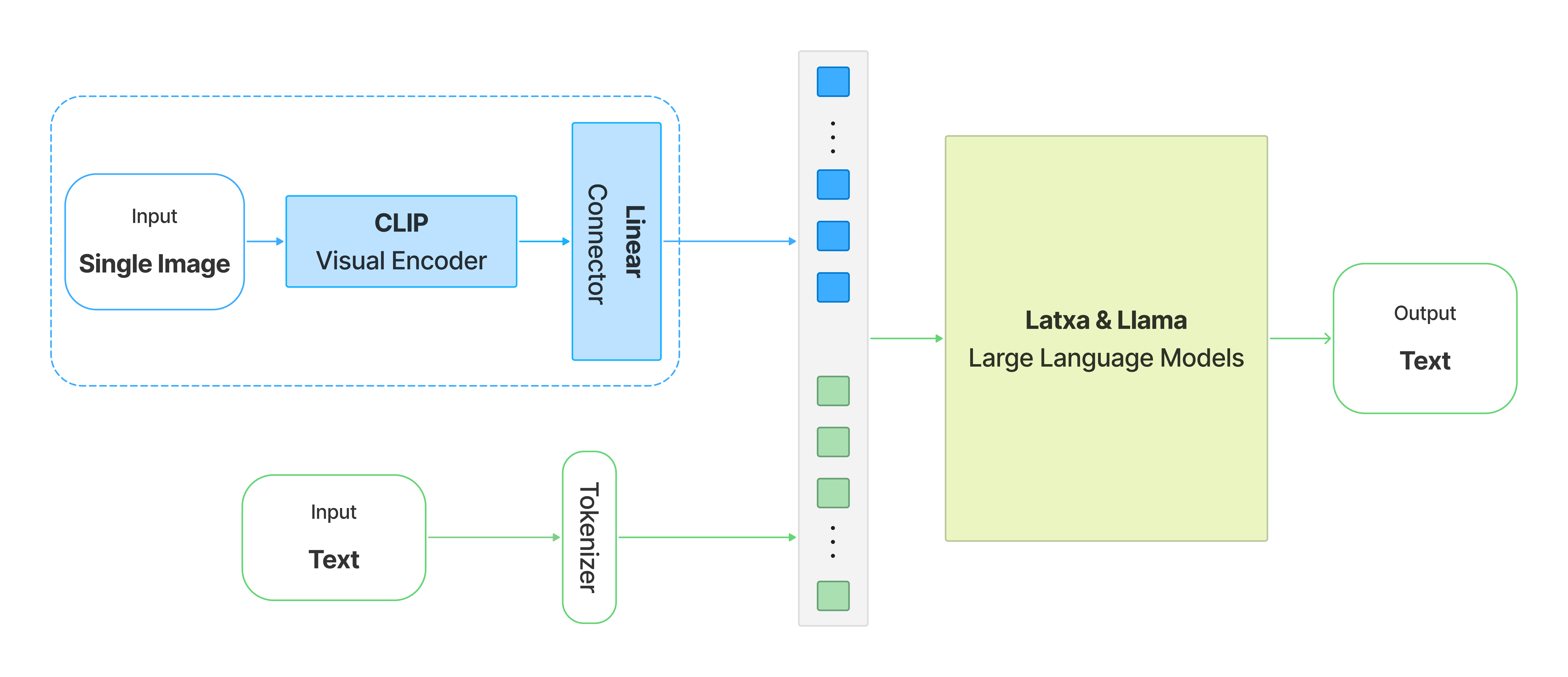}
    \caption{The late-fusion MLLM architecture used in this work. Our MLLMs have a visual encoder to represent input images, a connector to project visual representations into the embedding space of the LLM, and an LLM to process image inputs and textual queries to generate textual answers.}
    \label{fig:architecture}
\end{figure*}








\section{Related Work}
\label{sec:sota}


The current Multimodal Large Language Model field is dominated by proprietary models. According to the vision LMArena ranking\footnote{Accessed at: Oct 6th, 2025.} \cite{lmarena}, the leading models are Gemini-2.5-Pro \cite{geminipro}, GPT-5 \cite{gpt5}, and Claude-4 \cite{claude}. While these state-of-the-art MLLMs demonstrate superior performance, their architectures and training methodologies are largely unknown.

Among open weight alternatives, Qwen3-VL-235B \cite{qwen} ranks competitively in the arena leaderboard, achieving notable results. However, the composition of its dataset is still unknown. The Molmo-72B \cite{pixmo} is the best-performing model that offers open training data and is located around the 57th position in multimodal performance.

The vision LMArena and most other multimodal benchmarks primarily focus on evaluating the English performance of MLLMs, ignoring their multilingual performance. Consequently, efforts to develop open multilingual MLLMs are limited to a few works, such as Pangea \cite{pangea} and Ayavision \cite{ayavision}, which address this perspective by training on multimodal training data of multiple languages obtained through translation.

Despite these efforts, there is still a substantial performance gap between open and proprietary MLLMs \cite{pangea}. In fact, this gap is likely even more pronounced for low-resource languages, which are excluded from the training and evaluation datasets of existing works. To the best of our knowledge, this work represents the first effort at scale to develop an MLLM specifically for a low-resource language.

For Basque, this performance gap has recently been overcome in the text-only setting by Latxa-Llama-3.1 \cite{latxa}, a family of open LLMs based on Llama-3.1 \cite{llama} specially trained for Basque. These models have highlighted the potential of using translation for text-only datasets, showing competitive results with proprietary models.

\section{MLLM Architecture and Training}
\label{sec:mllm}

In this work, we adopt the late-fusion architecture for building MLLMs. This choice is motivated by two main factors: (i) the late-fusion design is prevalent among most open MLLMs \cite{cambrian, llava1.5, nvlm, pixmo}, and (ii) it enables us to assess the feasibility of already pretrained LLMs, such as Latxa. The following sections provide a detailed description of the architecture (\S\ref{sec:architecture}) and the two-stage training procedure typically used for developing such models (\S\ref{sec:training-procedure}).

\subsection{Architecture}
\label{sec:architecture}

The late-fusion architecture trains an instructed backbone LLM with the visual representations generated by a pretrained visual encoder (see Figure \ref{fig:architecture}). The role of this connector is to transform the visual representations into the embedding space of the LLM. Once converted, the embedding representations of both the visual features and the text tokens are processed simultaneously by the backbone LLM to produce the textual output required for a multimodal task. 

As the pretrained visual encoder, we have opted to use a CLIP visual encoder \cite{radford2021learning}, after performing some initial experiments with SigLIP \cite{zhai2023sigmoid} and obtaining very similar performances as in \cite{pixmo}. More concretely, we use the \texttt{clip-vit-large-patch14-336} ViT, which has already proven successful for the multilingual \cite{pangea} and general-purpose scenarios \cite{llava}.

Regarding the neural vision-language connector, we use a fully connected linear layer. This choice is based on PaliGemma \cite{paligemma}, which concludes that the single-layer linear connector outperforms more complex methods while being simpler and more efficient.

Finally, we compare the MLLMs based on the \texttt{Llama-3.1-8B-Instruct} (Llama) and \texttt{Latxa-Llama-3.1-8B-Instruct} (Latxa) models to assess the impact of a Basque-instructed backbone LLM. Since these LLMs share the same architecture and pretraining stage, they provide a fair comparison for evaluating how Basque-specific training influences posterior multimodal capabilities.


\subsection{Training procedure}
\label{sec:training-procedure}

Late-fusion MLLMs are usually trained following a two-stage procedure \cite{MLLMReview} consisting of : i) Vision-Language Alignment and ii) Multimodal Instruction Tuning.

\paragraph{Stage 1: Vision-Language Alignment.}
During the Vision–Language Alignment stage, the aim is to align the embedding spaces of the vision encoder and the backbone LLM. For this purpose, only the connector between them is trained while the other components remain frozen. This setup prevents the LLM from receiving out-of-distribution visual tokens in the second stage, thereby reducing data drift and stabilizing training.

We perform this stage equally for two backbone LLMs, namely Latxa and Llama, sharing the same training data and visual encoder. These systems will then be used for the subsequent Multimodal Instruction Tuning stage.

\paragraph{Stage 2: Multimodal Instruction Tuning.}

The Multimodal Instruction Tuning stage fine-tunes both the connector and the backbone LLM. This stage is where the backbone LLM learns to follow complex multimodal instructions. We experiment with various training configurations during the second phase to directly evaluate how different training dataset mixes affect the model's ability to process and generate responses.


\section{Basque Multimodal Datasets}
\begin{table*}[t!]
\centering
\begingroup
\renewcommand{\arraystretch}{1.5} 
\begin{tabular}{lcc}
\toprule
\textbf{Dataset} & \textbf{Acceptance Rate} & \textbf{Agreement} \\  
\midrule
\rowcolor{gray!20}
VQAv2\textsubscript{Eus} & 0.8375 & 0.887 \\
Pixmo-CapQA\textsubscript{Eus} & 0.8625 & 0.887 \\                                  
\rowcolor{gray!20}
A-OKVQA\textsubscript{Eus}  & 0.975  & 0.95 \\                                  
\bottomrule
\end{tabular}
\endgroup
\caption{Sample acceptance rate and mean inter-annotator agreement between the 4 annotators. The agreement has been calculated by using 40 images.}
\label{eval:eval_data_annotation}
\end{table*}

\label{sec:dataset}
Developing an MLLM for a specific language requires large-scale training datasets and evaluation benchmarks in the target language. Given the scarcity of resources for Basque, and following common practices in multilingual MLLMs \cite{pangea, ayavision}, we have created new multimodal datasets for Basque by translating from English-centric datasets. To do so, we analyzed the textual parts of each dataset to assess the most suitable translation method, alternating between sentence-level neural translators for Basque and text-only LLMs. The details for those procedures can be found in Appendix \ref{sec:Appendix1}. In total, we have created two multimodal datasets for training (\S\ref{sec:training-datasets}) and four evaluation benchmarks (\S\ref{sec:benchmarks}).


\subsection{Training datasets}
\label{sec:training-datasets}

\paragraph{Stage 1.} In the standard two-stage training procedure (\S\ref{sec:training-procedure}), image captioning is normally used for Stage 1, as it is a suitable task to align visual embeddings to the LLM embedding space. Following previous works \cite{llava, cambrian}, we opt to use the Conceptual Captions dataset (CC3M) \cite{cc3m}. CC3M consists of 3.3 million image-caption pairs, of which 2.8 million samples are available (some links are currently broken). Unlike other better curated datasets such as COCO \cite{coco}, this dataset sources image-text pairs directly from web content, resulting in significantly greater diversity. As the original captions of the dataset are mainly short sentences, we translate them into Basque using a sentence-level specialized translator system\footnote{https://huggingface.co/HiTZ/mt-hitz-en-eu}. The result is the CC3M\textsubscript{Eus} dataset, with a total of 2.8 million image captions in Basque.


\paragraph{Stage 2.} For the Multimodal Instruction Tuning stage, the models are typically trained on a set of diverse and carefully filtered multimodal instruction datasets. These datasets involve both general-purpose multimodal instructions and specialized datasets for specific multimodal tasks (OCR, document understanding, object counting, and so on). Since we have focused only on general-purpose capabilities, we have selected the Pixmo-ask-model-anything dataset (Pixmo-AMA) \cite{pixmo}. Pixmo-AMA contains a diverse collection of 162k human-annotated question–answer multimodal instructions. The dataset was generated through an iterative process in which a text-only LLM proposed multimodal instructions that were then evaluated by human annotators. Annotators could accept or reject each sample, and if rejected, they provided feedback to help the LLM refine its output until an acceptable answer was achieved. Given the complex questions and long answers of the dataset, requiring multi-sentence coherence, we translate it into Basque using the best open LLM possible, Latxa-Llama-3.1-Instruct-70B \cite{latxa}. Sentence-level translators were discarded since they do not keep the coherence of multi-sentence answers. In total, due to some of its images not being currently available, the Pixmo-AMA\textsubscript{Eus} dataset contains 146k images, questions and answers in Basque.

\subsection{Evaluation datasets}
\label{sec:benchmarks}
Following standard practices in the field \cite{nvlm, pixmo}, we evaluate our MLLMs for: i) multimodal understanding and ii) language generation with multimodal inputs. For evaluating multimodal understanding, we use close-ended benchmarks (\S\ref{sec:close-ended}), and for language generation, we build an open-ended benchmark (\S\ref{sec:open-ended}).

\subsubsection{Close-ended benchmarks}
\label{sec:close-ended}

As we are mainly interested in measuring the capabilities of Basque MLLMs in general multimodal understanding, we discard benchmarks that focus on specific skills such as OCR or table/chart understanding. In this context, we select three close-ended benchmarks: VQAv2 \cite{VQAv2}, A-OKVQA \cite{aokvqa} and Pixmo-CapQA \cite{pixmo}. 

\paragraph{VQAv2 \cite{VQAv2}} is the second iteration of the Visual Question-Answering dataset. The dataset focuses on the visual question-answering task, where given an image and a question, models have to answer that question. As opposed to the other close-ended evaluation benchmarks, VQAv2 requires generating single-word or short-phrase answers. Each instance contains a list of 10 human-generated answers as ground-truth. We use the VQA accuracy as the evaluation metric, as proposed by \citet{VQAv2}. However, since the MLLMs in our study were not specifically trained on specialized datasets with short answers, they cannot produce single-word answers even when implicitly requested in the prompt \cite{llava1.5}. To address this limitation, we modified the evaluation procedure of the benchmark to use inclusion rather than exact string matching. Under this approach, responses are considered correct as long as they contain a word from the ground truth answer anywhere within the generated text.

\paragraph{A-OKVQA \cite{aokvqa}} is a multiple-choice multimodal evaluation benchmark. It consists of 25k image-question pairs along with four possible answers to the provided question. We chose this benchmark because, in contrast with other VQA multiple-choice datasets, models need to relate their internal world knowledge with the visual input to answer the provided questions successfully. Following standard practice for multiple-choice benchmarks, the selected answer corresponds to the option with the highest log-probability at inference. Accuracy was then used as the evaluation metric, since the answer distribution is balanced. 
\begin{figure*}[t!]
\centering
\begin{tikzpicture}
\begin{axis}[
hide axis,
xmin=0, xmax=1,
ymin=0, ymax=1,
width=2cm,
height=2.5cm,
legend style={
at={(0.5,0.5)},
anchor=center,
legend columns=2,
/tikz/every even column/.append style={column sep=1cm}
},
]`
\addlegendimage{line width=1.5pt, color=red!60, mark=, mark options={fill=red!60, scale=0.8}}
\addlegendentry{English}
\addlegendimage{line width=1.5pt, color=blue!60, mark=, mark options={fill=blue!60, scale=0.8}}
\addlegendentry{Basque (translated)}
\end{axis}
\end{tikzpicture}
\vspace{0.3cm}

\begin{subfigure}[b]{0.32\textwidth}
\centering
\begin{tikzpicture}
\begin{axis}[
    xmin=-5, xmax=105,
    ymin=0, ymax=1.0,
    xtick={0, 20, 40, 60, 80, 100},
    ytick={0, 0.2, 0.4, 0.6, 0.8, 1.0},
    grid=major,
    width=5.5cm,
    height=4.0cm,
    ylabel={Performance},
    ylabel near ticks,
    title={VQAv2},
    title style={at={(0.5,0.95)}, anchor=south},
]
\addplot[line width=1.5pt, color=red!60, mark=*, mark options={fill=red!60, scale=0.8}] coordinates {
    (0, 0.62) (20, 0.63) (80, 0.61) (100, 0.05)
};
\addplot[line width=1.5pt, color=blue!60, mark=*, mark options={fill=blue!60, scale=0.8}] coordinates {
    (0, 0.45) (20, 0.59) (80, 0.62) (100, 0.61)
};
\end{axis}
\end{tikzpicture}
\end{subfigure}
\hfill
\begin{subfigure}[b]{0.32\textwidth}
\centering
\begin{tikzpicture}
\begin{axis}[
    xmin=-5, xmax=105,
    ymin=0, ymax=1.0,
    xtick={0, 20, 40, 60, 80, 100},
    ytick={0, 0.2, 0.4, 0.6, 0.8, 1.0},
    grid=major,
    width=5.5cm,
    height=4.0cm,
    title={A-OKVQA},
    title style={at={(0.5,0.95)}, anchor=south},
]
\addplot[line width=1.5pt, color=red!60, mark=*, mark options={fill=red!60, scale=0.8}] coordinates {
    (0, 0.59) (20, 0.63) (80, 0.63) (100, 0.59)
};
\addplot[line width=1.5pt, color=blue!60, mark=*, mark options={fill=blue!60, scale=0.8}] coordinates {
    (0, 0.50) (20, 0.55) (80, 0.59) (100, 0.52)
};
\end{axis}
\end{tikzpicture}
\end{subfigure}
\hfill
\begin{subfigure}[b]{0.32\textwidth}
\centering
\begin{tikzpicture}
\begin{axis}[
    xmin=-5, xmax=105,
    ymin=0, ymax=1.0,
    xtick={0, 20, 40, 60, 80, 100},
    ytick={0, 0.2, 0.4, 0.6, 0.8, 1.0},
    grid=major,
    width=5.5cm,
    height=4.0cm,
    title={PixMoCapQA},
    title style={at={(0.5,0.95)}, anchor=south},
]
\addplot[line width=1.5pt, color=red!60, mark=*, mark options={fill=red!60, scale=0.8}] coordinates {
    (0, 0.66) (20, 0.69) (80, 0.69) (100, 0.70)
};
\addplot[line width=1.5pt, color=blue!60, mark=*, mark options={fill=blue!60, scale=0.8}] coordinates {
    (0, 0.51) (20, 0.54) (80, 0.55) (100, 0.51)
};
\end{axis}
\end{tikzpicture}
\end{subfigure}
\begin{tikzpicture}[overlay, remember picture]
\fill[purple!35!white] (0.1, 0.5) rectangle (0.7, 2.9);
\node[rotate=90, anchor=center] at (0.4, 1.8) {\large \textbf{Latxa}};
\end{tikzpicture}

\vspace{0.7cm}

\begin{subfigure}[b]{0.32\textwidth}
\centering
\begin{tikzpicture}
\begin{axis}[
    xlabel={Basque Multimodal Data \%},
    xmin=-5, xmax=105,
    ymin=0, ymax=1.0,
    xtick={0, 20, 40, 60, 80, 100},
    ytick={0, 0.2, 0.4, 0.6, 0.8, 1.0},
    grid=major,
    width=5.5cm,
    height=4.0cm,
    xlabel near ticks,
    ylabel={Performance},
    ylabel near ticks,
]
\addplot[line width=1.5pt, color=red!60, mark=*, mark options={fill=red!60, scale=0.8}] coordinates {
    (0, 0.62) (20, 0.63) (80, 0.58) (100, 0.08)
};
\addplot[line width=1.5pt, color=blue!60, mark=*, mark options={fill=blue!60, scale=0.8}] coordinates {
    (0, 0.24) (20, 0.56) (80, 0.59) (100, 0.59)
};
\end{axis}
\end{tikzpicture}
\end{subfigure}
\hfill
\begin{subfigure}[b]{0.32\textwidth}
\centering
\begin{tikzpicture}
\begin{axis}[
    xlabel={Basque Multimodal Data \%},
    xmin=-5, xmax=105,
    ymin=0, ymax=1.0,
    xtick={0, 20, 40, 60, 80, 100},
    ytick={0, 0.2, 0.4, 0.6, 0.8, 1.0},
    grid=major,
    width=5.5cm,
    height=4.0cm,
    xlabel near ticks,
]
\addplot[line width=1.5pt, color=red!60, mark=*, mark options={fill=red!60, scale=0.8}] coordinates {
    (0, 0.54) (20, 0.63) (80, 0.65) (100, 0.62)
};
\addplot[line width=1.5pt, color=blue!60, mark=*, mark options={fill=blue!60, scale=0.8}] coordinates {
    (0, 0.35) (20, 0.51) (80, 0.58) (100, 0.58)
};
\end{axis}
\end{tikzpicture}
\end{subfigure}
\hfill
\begin{subfigure}[b]{0.32\textwidth}
\centering
\begin{tikzpicture}
\begin{axis}[
    xlabel={Basque Multimodal Data \%},
    xmin=-5, xmax=105,
    ymin=0, ymax=1.0,
    xtick={0, 20, 40, 60, 80, 100},
    ytick={0, 0.2, 0.4, 0.6, 0.8, 1.0},
    grid=major,
    width=5.5cm,
    height=4.0cm,
    xlabel near ticks,
]
\addplot[line width=1.5pt, color=red!60, mark=*, mark options={fill=red!60, scale=0.8}] coordinates {
    (0, 0.72) (20, 0.69) (80, 0.71) (100, 0.70)
};
\addplot[line width=1.5pt, color=blue!60, mark=*, mark options={fill=blue!60, scale=0.8}] coordinates {
    (0, 0.51) (20, 0.54) (80, 0.52) (100, 0.61)
};
\end{axis}
\end{tikzpicture}
\end{subfigure}
\begin{tikzpicture}[overlay, remember picture]
\fill[blue!35!white] (0.1, 1.1) rectangle (0.7, 3.5);
\node[rotate=90, anchor=center] at (0.4, 2.3) {\large \textbf{Llama}};
\end{tikzpicture}

\caption{Accuracy across multimodal benchmarks of Latxa-based (top row) and Llama-based (bottom row) MLLMs trained with different percentages of Basque Multimodal Instruction Data. The models are evaluated on the English (original) and Basque (translated) versions of close-ended benchmarks. }
\label{fig:performance_benchmark_multimodal}
\end{figure*}

\paragraph{PixMo-CapQA \cite{pixmo}} is a semi-automatically generated multimodal dataset that derives from dense image captions. It consists of 214k question-answer pairs generated from 165k distinct images, from which we have filtered only the yes/no questions for evaluation purposes. Therefore, we have created a multimodal binary multiple-choice dataset that requires models to demonstrate knowledge comprehension to answer the provided questions successfully. We follow the same evaluation procedure as in A-OKVQA, using the log-probabilities of the models to select the answer and accuracy as the evaluation metric.

For each close-ended benchmark, we have created a subset of 2.5k randomly sampled examples. We translate the three benchmarks with the Latxa-Llama-3.1-Instruct 70B \cite{latxa}, feeding the model with the original question and all possible answers for a given instance. We created specific translation prompts for each benchmark, using two-shot examples to follow a specific format. The specific prompts will be found in Appendix \ref{sec:Appendix1}. As a result of this translation process, we created the Basque evaluation benchmarks VQAv2\textsubscript{Eus}, A-OKVQA\textsubscript{Eus}, and Pixmo-CapQA\textsubscript{Eus}, and their English parallel versions. In this way, we can evaluate MLLMs both in English and Basque using the same instances. Evaluations were conducted using the LLMs-eval \cite{lmmseval} suite. We extended its support to include the \textit{A-OKVQA} and \textit{PixMo-CapQA} datasets in addition to the already supported \textit{VQAv2}. 

The quality of the translations is crucial for evaluation benchmarks, so four Basque native speakers validated our benchmarks. For each benchmark, samples were uniformly sampled for annotation. Each annotator evaluated 40 question–answer pairs per benchmark. From which 20 are from a shared common set and 20 from their own independent set. Across all annotators, each benchmark received a total of 100 annotations. The inter-annotator agreement is the mean of accuracy between the four annotators on each image of the shared common set. Table \ref{eval:eval_data_annotation} shows high acceptance rates and agreement, confirming the quality of the translations and the suitability of our benchmarks for MLLM evaluation.

\subsubsection{Open-ended benchmarks}
\label{sec:open-ended}
Close-ended benchmarks are a good option to assess the multimodal understanding of models. However, we also want to measure the quality and coherence of the generated textual output for multimodal scenarios. Thus, we also use an open-ended generation benchmark.

\paragraph{WildVision \cite{wildvision}} is a recent benchmark consisting of 500 text-image pairs obtained from the WildVision-Arena; a platform that collected human preferences to evaluate MLLMs. Due to numerous samples requiring specific multimodal capabilities such as OCR, a human annotator has filtered the dataset to only incorporate those pairs based on general multimodal capabilities. The filtered dataset resulted in a total of 199 samples. We translated those samples to Basque using Latxa-Llama-3.1-Instruct 70B \cite{latxa} and manually review the quality of all the samples. As a result, we created the WildVision\textsubscript{Eus} dataset, with 199 images and open-ended questions in Basque. Recall that the original benchmark does not provide any ground-truth answers, which complicates the evaluation process (\S\ref{sec:experiments}).

\begin{figure*}[t!]
\centering
\begin{tikzpicture}
\begin{axis}[
hide axis,
xmin=0, xmax=1,
ymin=0, ymax=1,
width=10cm,
height=2.5cm,
legend style={
at={(0.5,0.5)},
anchor=center,
legend columns=3,
/tikz/every even column/.append style={column sep=1cm}
},
]
\addlegendimage{line width=1.5pt, color=teal!85, mark=, mark options={fill=teal!85, scale=0.8}]}
\addlegendentry{Multimodal}

\addlegendimage{line width=1.5pt, color=purple!75, mark=, mark options={fill=purple!75, scale=0.8}}
\addlegendentry{Multimodal + Text-only}

\addlegendimage{dashed, line width=1.5pt, color=black!}
\addlegendentry{Base LLM performance}

\end{axis}
\end{tikzpicture}
\vspace{0.3cm}

\begin{subfigure}[b]{0.45\textwidth}
\centering
\begin{tikzpicture}
\begin{axis}[
    xmin=-5, xmax=105,
    ymin=0, ymax=1.0,
    xtick={0, 20, 40, 60, 80, 100},
    ytick={0, 0.2, 0.4, 0.6, 0.8, 1.0},
    grid=major,
    width=7cm,
    height=4.0cm,
    ylabel={Performance},
    ylabel near ticks,
    title={Multimodal performance (VQAv2\textsubscript{Eus})},
    title style={at={(0.5,0.95)}, anchor=south},
]
\addplot[line width=1.5pt, color=teal!85, mark=*, mark options={fill=teal!85, scale=0.8}] coordinates {
    (0, 0.45) (20, 0.59) (80, 0.62) (100, 0.61)
};
\addplot[line width=1.5pt, color=purple!75, mark=*, mark options={fill=purple!75, scale=0.8}] coordinates {
    (0, 0.39) (20, 0.59) (80, 0.61) (100, 0.56)
};
\end{axis}
\end{tikzpicture}
\end{subfigure}
\hfill
\begin{subfigure}[b]{0.45\textwidth}
\centering
\begin{tikzpicture}
\begin{axis}[
    xmin=-5, xmax=105,
    ymin=0, ymax=1.0,
    xtick={0, 20, 40, 60, 80, 100},
    ytick={0, 0.2, 0.4, 0.6, 0.8, 1.0},
    grid=major,
    width=7cm,
    height=4.0cm,
    title={Text-only performance (BertaQA local)},
    title style={at={(0.5,0.95)}, anchor=south},
]
\addplot[dashed, line width=2pt, color=black!70] coordinates {(0, 0.66) (100, 0.66)};
\addplot[line width=1.5pt, color=teal!85, mark=*, mark options={fill=teal!85, scale=0.8}] coordinates {
    (0, 0.52) (20, 0.51) (80, 0.52) (100, 0.53)

};
\addplot[line width=1.5pt, color=purple!75, mark=*, mark options={fill=purple!75, scale=0.8}] coordinates {
    (0, 0.60) (20, 0.55) (80, 0.53) (100, 0.61)
};
\end{axis}
\end{tikzpicture}
\end{subfigure}
\begin{tikzpicture}[overlay, remember picture]
\fill[purple!35!white] (0.1, 0.5) rectangle (0.7, 2.9);
\node[rotate=90, anchor=center] at (0.4, 1.7) {\large \textbf{Latxa}};
\end{tikzpicture}
\vspace{0.7cm}

\begin{subfigure}[b]{0.45\textwidth}
\centering
\begin{tikzpicture}
\begin{axis}[
    xlabel={Basque Multimodal Data \%},
    xmin=-5, xmax=105,
    ymin=0, ymax=1.0,
    xtick={0, 20, 40, 60, 80, 100},
    ytick={0, 0.2, 0.4, 0.6, 0.8, 1.0},
    grid=major,
    width=7cm,
    height=4.0cm,
    xlabel near ticks,
    ylabel={Performance},
    ylabel near ticks,
    title style={at={(0.5,0.95)}, anchor=south},
]
\addplot[line width=1.5pt, color=teal!85, mark=*, mark options={fill=teal!85, scale=0.8}] coordinates {
    (0, 0.24) (20, 0.56) (80, 0.59) (100, 0.59)
};
\addplot[line width=1.5pt, color=purple!75, mark=*, mark options={fill=purple!75, scale=0.8}] coordinates {
    (0, 0.50) (20, 0.57) (80, 0.61) (100, 0.60)
};
\end{axis}
\end{tikzpicture}
\end{subfigure}
\hfill
\begin{subfigure}[b]{0.45\textwidth}
\centering
\begin{tikzpicture}
\begin{axis}[
    xlabel={Basque Multimodal Data \%},
    xmin=-5, xmax=105,
    ymin=0, ymax=1.0,
    xtick={0, 20, 40, 60, 80, 100},
    ytick={0, 0.2, 0.4, 0.6, 0.8, 1.0},
    grid=major,
    width=7cm,
    height=4.0cm,
    xlabel near ticks,
]
\addplot[dashed, line width=2pt, color=black!70] coordinates {(0, 0.45) (100, 0.45)};
\addplot[line width=1.5pt, color=teal!85, mark=*, mark options={fill=teal!85, scale=0.8}] coordinates {
    (0, 0.40) (20, 0.39) (80, 0.41) (100, 0.38)
};
\addplot[line width=1.5pt, color=purple!75, mark=*, mark options={fill=purple!75, scale=0.8}] coordinates {
    (0, 0.41) (20, 0.43) (80, 0.43) (100, 0.41)
};
\end{axis}
\end{tikzpicture}
\end{subfigure}
\begin{tikzpicture}[overlay, remember picture]
\fill[blue!35!white] (0.1, 1.1) rectangle (0.7, 3.5);
\node[rotate=90, anchor=center] at (0.4, 2.3) {\large \textbf{Llama}};
\end{tikzpicture}

\caption{Accuracy across Basque multimodal (left) and text-only (right) benchmarks of Latxa-based (top row) and Llama-based (bottom row) MLLMs  trained with different percentages of Basque Multimodal Instruction Data. The models are evaluated on the English (original) and Basque (translated) versions.}
\label{fig:performance_benchmark_text_only}
\end{figure*}

\section{Experiments and Findings}

\label{sec:experiments}
We evaluate our Basque MLLMs for close-ended and open-ended generation tasks, both in English and Basque. To train our models, we have built upon the training codebase of Llava \cite{llava},  following the recommended hyperparameters by Pangea \cite{pangea}. The codebase will be public, and the specifics of the training configuration and infrastructure will be explained in detail in Appendix \ref{sec:Appendix2}.

All MLLM configurations share the same Stage 1 training recipe and data. We trained the two Stage 1 models in a mix of the translated and original CC3M dataset. However, due to the lack of previous studies on the optimal proportions of multilingual data for this stage, we followed the recommendations for Stage 2 \cite{pangea} while focusing on the Basque language. Particularly, we have chosen a ratio of 80\% of Basque and 20\% English samples. As for the hyperparameters, we used the AdamW optimizer together with a cosine learning rate scheduler with an initial learning rate of $10^{-3}$ and a warm-up ratio of 0.03. We use mixed-precision training and a sequence length of 8192 to achieve a global batch size of 32 samples.

For Stage 2, we have used the Pixmo-AMA dataset to explore data mixture strategies for multimodal instruction tuning (\S\ref{sec:multimodalStrategies}), how text-only performance behaves compared to the original backbone LLM (\S\ref{sec:text-only}), and whether a Basque instructed LLM is required for a strong Basque MLLM (\S\ref{sec:latxa-vs-llama}). Contrary to Stage 1, we initialized the learning rate scheduler at $2 \times 10^{-5}$, with a weight decay of 0.01 and a warm-up ratio of 0.03. We use mixed-precision training and a sequence length of 8192 to achieve a global batch size of 128 samples.

\begin{table*}[t!]
\centering
\small
\setlength{\tabcolsep}{10pt}
\renewcommand{\arraystretch}{1.3}
\begin{tabular}{@{}lcccccc@{}}
\toprule
\multirow{2}{*}{} & \multicolumn{3}{c}{\textbf{Human}} & \multicolumn{3}{c}{\textbf{GPT-5}} \\
\cmidrule(lr){2-4} \cmidrule(lr){5-7}
& \textbf{Latxa} & \textbf{Tie} & \textbf{Llama} & \textbf{Latxa} & \textbf{Tie} & \textbf{Llama} \\
\midrule
\textbf{Winning \%} & 23.11 & 53.79 & 20.45 & 31.44 & 36.74 & 31.82· \\
\bottomrule
\end{tabular}
\caption{Multimodal open-ended performance comparison between top-performing Llama-based and Latxa-based MLLMs across human and automated evaluation paradigms. Human evaluations resulted in inconclusive outcomes for 3\% of samples due to annotation discrepancies.}
\label{tb:human_annotation}
\end{table*}

\subsection{Exploring multimodal data mixture strategies}
\label{sec:multimodalStrategies}
To study how the training data mixtures affect the MLLM performance, we trained Latxa-based and Llama-based architectures on the Pixmo-AMA dataset using four Basque–English sample ratios: 0:100, 20:80, 80:20, and 100:0. It should be noted that each split contains the same underlying samples, with only their languages being swapped between English and Basque.

We train a total of eight configurations (2 architectures on 4 training splits), which we evaluate using the three multimodal close-ended benchmarks described in Section \ref{sec:dataset}: VQAv2, A-OKVQA and PixMoCapQA, both in English and Basque. The results are presented in Figure \ref{fig:performance_benchmark_multimodal}. As can be observed, the best configurations for both backbone LLMs coincide using 80\% Basque data and 20\% English. Furthermore, Latxa and Llama backbone LLMs, obtain a similar average accuracy of 0.62 and 0.61 points across the three benchmarks in both languages, respectively. This has been the highest average among the four data-mixture ratios, although with very low margins.

\paragraph{Finding 1: Low ratios of Basque multimodal data are enough to obtain solid results in Basque benchmarks.} In fact, the performance of the models in Basque remains comparable when at least 20\% of Basque multimodal data is used to train the model. In VQAv2, which is the most informative among the three benchmarks, the Llama-based configuration trained on the split composed of 100\% Basque multimodal data surpasses the version trained with only 20\% by only 0.03 absolute points. This gain is even smaller for the Latxa-based configuration, whose improvement is only 0.02 points. Our finding is also supported by the Pangea study \cite{pangea}, which reports a similar trend where performance in multimodal tasks reached a plateau once the proportion of multilingual data exceeded 20\%.

\paragraph{Finding 2: A low ratio of English data is required to avoid catastrophic forgetting in English benchmarks.} As expressed in the VQAv2 benchmark, both backbone LLM architectures show a drastic decline in English performance when only Basque multimodal data is used for training, indicating catastrophic forgetting of their English capabilities. These results highlight the need for English multimodal samples at training time to maintain English performance.



\subsection{Evaluating text-only performance}
\label{sec:text-only}
Multiple studies have reported that multimodal training tends to reduce the performance of backbone LLMs on text-only tasks \cite{paligemma, pangea}. Incorporating text-only instructions into the training process has proven to be an effective way to counteract this decline \cite{nvlm}. To analyze this effect in a low-resource context: i) we augmented the four multimodal training splits described in Section \ref{sec:multimodalStrategies} with text-only instructions, ii) we trained MLLMs using those new training configurations, and iii) we evaluated the resulting models on both multimodal and text-only benchmarks.

The text-only instructions were randomly sampled from the Magpie-Llama-3.1-8B-Instruct-Filtered-1M dataset, originally developed for training Latxa \cite{latxa}. We added a total of 29k text-only examples, 20\% of them being in English and 80\% in Basque while maintaining the multimodal samples. Consequently, the final training datasets consist of 173k samples, from which 17\% is text-only and 83\% is multimodal data.

In addition to multimodal evaluation, we evaluate the models for text-only capabilities using the BertaQA \cite{bertaqa} benchmark, a Basque multiple-choice trivia dataset emphasizing local cultural knowledge. This benchmark has been chosen for its strong correlation with human evaluation, as noted in \cite{latxa}.

Figure \ref{fig:performance_benchmark_text_only} presents the performance of our previous eight configurations, comparing them with the addition of the text-only training split to the multimodal one. In this case, the text-only training split always uses a fixed 80-20 Basque-English ratio, whereas the multimodal splits' ratios are changed according to these configurations. We report only multimodal results for VQAv2\textsubscript{Eus}, as we find this benchmark to best represent the overall Basque multimodal performance.

\paragraph{Finding 3: Incorporating text-only data helps reduce the Basque text-only performance drop observed relative to the original backbone LLM.} Although this addition leads to an overall improvement in text-only performance, the degradation compared to the base LLM remains considerable, especially in Latxa-based configurations. Notably, this degradation is smaller when the split consists exclusively of either English or Basque content. Given that previous studies, such as \cite{nvlm}, have demonstrated that this degradation can be effectively addressed by adding text-only data, our results may be caused by insufficient text-only instruction samples.



\paragraph{Finding 4: Text-only data can improve multimodal performance.} This is especially the case of the Llama-based configuration trained solely with English multimodal data. The inclusion of Basque text-only data has significantly improved its multimodal performance in Basque. This suggests that, in the absence of Basque multimodal data, Basque text-only instructions can provide a bridge to transfer the multimodal capabilities acquired in English to Basque. This finding indicates modality-transfer capabilities when no multimodal data is present in the language of evaluation. We find this to be of special interest to address the scarcity of multimodal data for low-resource languages.

\subsection{Evaluating the impact of backbone LLMs}
\label{sec:latxa-vs-llama}

The results of the close-ended evaluations, Figures \ref{fig:performance_benchmark_multimodal} and \ref{fig:performance_benchmark_text_only}, show similar performance between Latxa-based and Llama-based architectures in most of the configurations, suggesting that a Basque instructed backbone LLM has no advantage over a mostly English model. As close-ended generation may not demand language proficiency, we have evaluated our MLLMs in the Wildvision open-ended benchmark, to get a clearer view of the importance of the backbone LLM. 

We opt for human evaluation, and due to the low scalability of this procedure, we have only evaluated the best-performing Latxa-based and Llama-based configurations. That is, we evaluate the models trained with the ratio of 80-20 Basque-English multimodal instructions, in addition to the text-only dataset. 

For each question, the answers generated by both configurations have been compared pairwise to determine the preferred response or whether the result was a tie. The annotation process considered three evaluation aspects ordered by importance.

\begin{enumerate}
    \item \textbf{Relevance and language}: Whether the response is satisfactory and is written in Basque.
    \item \textbf{Quality}: Overall quality of the response in terms of correctness.
    \item \textbf{Language proficiency}: The proficiency of the response in the Basque language.
\end{enumerate}

For the annotation process, four native Basque speakers rated the 199 samples of the Basque Wildvision dataset (\S\ref{sec:dataset}). All annotators evaluated the same 44 responses to calculate the inter-annotator agreement, and we used majority voting to decide the final rating for those samples. Given the lack of scalability of this evaluation approach, we also conducted the same evaluation following an MLLM-as-a-judge paradigm. More concretely, we use the \texttt{GPT-5-2025-08-07} model, which is prompted with the question, the image, and the two responses generated by Llama-based and Latxa-based MLLMs. We also prompt GPT-5 with the set of evaluation criteria to guide its assessment.

The results obtained by both evaluation methods can be found in Table \ref{tb:human_annotation}. Notice that we will focus on human evaluation results for the analysis of the performance of MLLMs. The inter-annotator agreement among humans, measured as Cohen's Kappa, is of 0.43, which is considered as moderate.





\paragraph{Finding 5: A Basque backbone LLM is not required to develop a strong Basque MLLM.} In fact, Latxa-based and Llama-based configurations exhibited comparable performance in 54\% of the open-ended benchmarks, confirming the minimal differences observed between the two models in the close-ended benchmarks.  Although the Latxa backbone shows a slight performance advantage, the difference is not significant given the low number of test samples (199), leading us to conclude that both models perform similarly. 



\paragraph{Finding 6: Judge models may offer a good way to evaluate Basque open-ended generations, but we cannot prescind from human evaluation.} Although there is a fair agreement between GPT-5 and human annotators (Cohen's Kappa 0.33), GPT-5 tends to produce significantly fewer ties, even when explicitly instructed to do so in the prompt. This reflects a documented bias in LLM-based evaluators, as discussed in prior research \cite{mllmAsJudge}. This means that, at least for Basque, MLLM-as-a-judge is a promising method to evaluate open-ended generation at bigger scales, but as some differences with human annotators are still observed, it cannot be used in isolation yet. 


\section{Conclusions}

In this work, we create a total of 2 multimodal training datasets and 4 evaluation benchmarks for Basque, a low-resource language. We use these resources develop the first Multimodal Large Language Model for Basque, exploring several training strategies. We find that low ratios of Basque multimodal data are already enough to perform well for Basque multimodal benchmarks. More importantly, we also find that the inclusion of Basque text-only data can enhance the multimodal performance of MLLMs in Basque, showcasing the cross-lingual transfer capabilities of these systems. Furthermore, we see that the performance of English-centric backbone LLMs are close to Basque-centric LLMs. Those three findings together indicate that training an English-centric LLM with minimal multimodal data for the target language combined by text-only instructions in that target language can provide a feasible pathway for developing MLLMs for many low-resource languages.



Many directions are open for future work. One of the important aspects that could be explored in the future is training and evaluating on multimodal cultural knowledge. As this work relies on machine translation for training and evaluation data, no Basque cultural knowledge is included. Another interesting direction is measuring how far we can get without using any multimodal data in the target language, by leveraging already trained MLLMs and cross-lingual and cross-modality transfer. Finally, more and better multimodal training and evaluation data are needed, especially for specific skills such as OCR or table/chart understanding. 

\section*{Limitations}

This work explores various approaches for building MLLMs for low-resource languages. However, due to the computational cost of experiments and the human annotations required, we had to limit the exploration.

One of the limitations is the choice of a single language. Although Basque is representative of many low-resource languages in the number of resources available, it could be the case that our results do not fully generalize to other low-resource languages.

Another limitation is our choice of architecture and model, where we focus on a single architecture and two different LLMs. This decision was motivated by the availability of Latxa. However, other approaches might indeed be promising, such as leveraging a powerful MLLM and training it in the target language.

Finally, due to the lack of resources available for Basque, we had to rely on machine translation for both training and evaluation. Although we validated the translated benchmarks with human annotations, this method has its inherent limitations \cite{translation}.

\section*{Ethics Statement}
MLLMs have the potential to impact society. While they can improve global information access and enable new forms of automation, they also present risks, including the enhancement of existing human biases and privacy issues for individuals. In this direction, the adoption of multilingual MLLMs for low-resource languages has the particular risk of extending this issue to local behaviors. Therefore, the development of these systems must prioritize responsible practices to address these challenges.

\section*{Acknowledgements}

This work has been partially supported by the Basque Government (Research group funding IT1570-22 and IKER-GAITU-2 project), the Spanish Ministry of Science, Innovation and Universities (Molvi project PID2024-157855OB-C32, HumanAIze project AIA2025-163322-C61 and  Geo-R2-LLM project PCI2025-163286 by MICIU/AEI /10.13039/501100011033 and co-financed by the European Union) and the European Union’s Horizon Europe research and innovation programme under Grant Agreement No 10113572, related to the LUMINOUS project. Julen Etxaniz holds a PhD grant from the Basque Government (PRE\_2024\_2\_0028).

The models were trained on the Leonardo supercomputer at CINECA under the EuroHPC Joint Undertaking, project EHPC-EXT-2024E01-042. The authors also acknowledge the technical and human support provided by the DIPC Supercomputing Center. 

\section*{Bibliographical References}\label{sec:reference}

\bibliographystyle{lrec2026-natbib}
\bibliography{bibliography}

\begin{thebibliography}{30}
\expandafter\ifx\csname natexlab\endcsname\relax\def\natexlab#1{#1}\fi

\bibitem[{Anthropic(2024)}]{claude}
Anthropic. 2024.
\newblock Claude opus 4.
\newblock \url{https://www.anthropic.com}.
\newblock Large language model.

\bibitem[{Artetxe et~al.(2020)Artetxe, Labaka, and Agirre}]{translation}
Mikel Artetxe, Gorka Labaka, and Eneko Agirre. 2020.
\newblock \href {https://doi.org/10.18653/v1/2020.emnlp-main.618} {Translation artifacts in cross-lingual transfer learning}.
\newblock In \emph{Proceedings of the 2020 Conference on Empirical Methods in Natural Language Processing (EMNLP)}, pages 7674--7684, Online. Association for Computational Linguistics.

\bibitem[{Beyer et~al.(2024)Beyer, Steiner, Pinto, Kolesnikov, Wang, Salz, Neumann, Alabdulmohsin, Tschannen, Bugliarello, Unterthiner, Keysers, Koppula, Liu, Grycner, Gritsenko, Houlsby, Kumar, Rong, Eisenschlos, Kabra, Bauer, Bošnjak, Chen, Minderer, Voigtlaender, Bica, Balazevic, Puigcerver, Papalampidi, Henaff, Xiong, Soricut, Harmsen, and Zhai}]{paligemma}
Lucas Beyer, Andreas Steiner, André~Susano Pinto, Alexander Kolesnikov, Xiao Wang, Daniel Salz, Maxim Neumann, Ibrahim Alabdulmohsin, Michael Tschannen, Emanuele Bugliarello, Thomas Unterthiner, Daniel Keysers, Skanda Koppula, Fangyu Liu, Adam Grycner, Alexey Gritsenko, Neil Houlsby, Manoj Kumar, Keran Rong, Julian Eisenschlos, Rishabh Kabra, Matthias Bauer, Matko Bošnjak, Xi~Chen, Matthias Minderer, Paul Voigtlaender, Ioana Bica, Ivana Balazevic, Joan Puigcerver, Pinelopi Papalampidi, Olivier Henaff, Xi~Xiong, Radu Soricut, Jeremiah Harmsen, and Xiaohua Zhai. 2024.
\newblock \href {http://arxiv.org/abs/2407.07726} {Paligemma: A versatile 3b vlm for transfer}.

\bibitem[{Chen et~al.(2024)Chen, Chen, Zhang, Wang, Liu, Zhou, Zhang, Wan, Zhou, and Sun}]{mllmAsJudge}
Dongping Chen, Ruoxi Chen, Shilin Zhang, Yaochen Wang, Yinuo Liu, Huichi Zhou, Qihui Zhang, Yao Wan, Pan Zhou, and Lichao Sun. 2024.
\newblock Mllm-as-a-judge: assessing multimodal llm-as-a-judge with vision-language benchmark.
\newblock In \emph{Proceedings of the 41st International Conference on Machine Learning}, ICML'24. JMLR.org.

\bibitem[{Chiang et~al.(2024)Chiang, Zheng, Sheng, Angelopoulos, Li, Li, Zhu, Zhang, Jordan, Gonzalez, and Stoica}]{lmarena}
Wei-Lin Chiang, Lianmin Zheng, Ying Sheng, Anastasios~N. Angelopoulos, Tianle Li, Dacheng Li, Banghua Zhu, Hao Zhang, Michael~I. Jordan, Joseph~E. Gonzalez, and Ion Stoica. 2024.
\newblock Chatbot arena: an open platform for evaluating llms by human preference.
\newblock In \emph{Proceedings of the 41st International Conference on Machine Learning}, ICML'24. JMLR.org.

\bibitem[{Dai et~al.(2024)Dai, Lee, Wang, Yang, Liu, Barker, Rintamaki, Shoeybi, Catanzaro, and Ping}]{nvlm}
Wenliang Dai, Nayeon Lee, Boxin Wang, Zhuolin Yang, Zihan Liu, Jon Barker, Tuomas Rintamaki, Mohammad Shoeybi, Bryan Catanzaro, and Wei Ping. 2024.
\newblock \href {http://arxiv.org/abs/2409.11402} {Nvlm: Open frontier-class multimodal llms}.

\bibitem[{Dash et~al.(2025)Dash, Nan, Dang, Ahmadian, Singh, Smith, Venkitesh, Shmyhlo, Aryabumi, Beller-Morales, Pekmez, Ozuzu, Richemond, Locatelli, Frosst, Blunsom, Gomez, Zhang, Fadaee, Govindassamy, Roy, Gallé, Ermis, Üstün, and Hooker}]{ayavision}
Saurabh Dash, Yiyang Nan, John Dang, Arash Ahmadian, Shivalika Singh, Madeline Smith, Bharat Venkitesh, Vlad Shmyhlo, Viraat Aryabumi, Walter Beller-Morales, Jeremy Pekmez, Jason Ozuzu, Pierre Richemond, Acyr Locatelli, Nick Frosst, Phil Blunsom, Aidan Gomez, Ivan Zhang, Marzieh Fadaee, Manoj Govindassamy, Sudip Roy, Matthias Gallé, Beyza Ermis, Ahmet Üstün, and Sara Hooker. 2025.
\newblock \href {http://arxiv.org/abs/2505.08751} {Aya vision: Advancing the frontier of multilingual multimodality}.

\bibitem[{DeepMind(2025)}]{geminipro}
Google DeepMind. 2025.
\newblock Gemini 2.5 pro.
\newblock \url{https://ai.google.dev/gemini-api/docs/models}.
\newblock Advanced reasoning model. Model ID: gemini-2.5-pro.

\bibitem[{Deitke et~al.(2025)Deitke, Clark, Lee, Tripathi, Yang, Park, Salehi, Muennighoff, Lo, Soldaini, Lu, Anderson, Bransom, Ehsani, Ngo, Chen, Patel, Yatskar, Callison-Burch, Head, Hendrix, Bastani, VanderBilt, Lambert, Chou, Chheda, Sparks, Skjonsberg, Schmitz, Sarnat, Bischoff, Walsh, Newell, Wolters, Gupta, Zeng, Borchardt, Groeneveld, Nam, Lebrecht, Wittlif, Schoenick, Michel, Krishna, Weihs, Smith, Hajishirzi, Girshick, Farhadi, and Kembhavi}]{pixmo}
Matt Deitke, Christopher Clark, Sangho Lee, Rohun Tripathi, Yue Yang, Jae~Sung Park, Mohammadreza Salehi, Niklas Muennighoff, Kyle Lo, Luca Soldaini, Jiasen Lu, Taira Anderson, Erin Bransom, Kiana Ehsani, Huong Ngo, YenSung Chen, Ajay Patel, Mark Yatskar, Chris Callison-Burch, Andrew Head, Rose Hendrix, Favyen Bastani, Eli VanderBilt, Nathan Lambert, Yvonne Chou, Arnavi Chheda, Jenna Sparks, Sam Skjonsberg, Michael Schmitz, Aaron Sarnat, Byron Bischoff, Pete Walsh, Chris Newell, Piper Wolters, Tanmay Gupta, Kuo-Hao Zeng, Jon Borchardt, Dirk Groeneveld, Crystal Nam, Sophie Lebrecht, Caitlin Wittlif, Carissa Schoenick, Oscar Michel, Ranjay Krishna, Luca Weihs, Noah~A. Smith, Hannaneh Hajishirzi, Ross Girshick, Ali Farhadi, and Aniruddha Kembhavi. 2025.
\newblock Molmo and pixmo: Open weights and open data for state-of-the-art vision-language models.
\newblock In \emph{Proceedings of the IEEE/CVF Conference on Computer Vision and Pattern Recognition (CVPR)}, pages 91--104.

\bibitem[{Etxaniz et~al.(2025)Etxaniz, Azkune, Soroa, de~Lacalle, and Artetxe}]{bertaqa}
Julen Etxaniz, Gorka Azkune, Aitor Soroa, Oier~Lopez de~Lacalle, and Mikel Artetxe. 2025.
\newblock Bertaqa: how much do language models know about local culture?
\newblock In \emph{Proceedings of the 38th International Conference on Neural Information Processing Systems}, NIPS '24, Red Hook, NY, USA. Curran Associates Inc.

\bibitem[{Goyal et~al.(2016)Goyal, Khot, Summers-Stay, Batra, and Parikh}]{VQAv2}
Yash Goyal, Tejas Khot, Douglas Summers-Stay, Dhruv Batra, and Devi Parikh. 2016.
\newblock \href {https://api.semanticscholar.org/CorpusID:8081284} {Making the v in vqa matter: Elevating the role of image understanding in visual question answering}.
\newblock \emph{International Journal of Computer Vision}, 127:398 -- 414.

\bibitem[{Lin et~al.(2014)Lin, Maire, Belongie, Hays, Perona, Ramanan, Doll{\'a}r, and Zitnick}]{coco}
Tsung-Yi Lin, Michael Maire, Serge~J. Belongie, James Hays, Pietro Perona, Deva Ramanan, Piotr Doll{\'a}r, and C.~Lawrence Zitnick. 2014.
\newblock \href {https://api.semanticscholar.org/CorpusID:14113767} {Microsoft coco: Common objects in context}.
\newblock In \emph{European Conference on Computer Vision}.

\bibitem[{Liu et~al.(2024)Liu, Li, Li, and Lee}]{llava1.5}
Haotian Liu, Chunyuan Li, Yuheng Li, and Yong~Jae Lee. 2024.
\newblock Improved baselines with visual instruction tuning.
\newblock In \emph{Proceedings of the IEEE/CVF Conference on Computer Vision and Pattern Recognition (CVPR)}, pages 26296--26306.

\bibitem[{Liu et~al.(2023)Liu, Li, Wu, and Lee}]{llava}
Haotian Liu, Chunyuan Li, Qingyang Wu, and Yong~Jae Lee. 2023.
\newblock Visual instruction tuning.

\bibitem[{Llama-Team(2024)}]{llama}
Llama-Team. 2024.
\newblock \href {http://arxiv.org/abs/2407.21783} {The llama 3 herd of models}.

\bibitem[{Loshchilov and Hutter(2019)}]{adamw}
Ilya Loshchilov and Frank Hutter. 2019.
\newblock \href {https://openreview.net/forum?id=Bkg6RiCqY7} {Decoupled weight decay regularization}.
\newblock In \emph{International Conference on Learning Representations}.

\bibitem[{Lu et~al.(2024)Lu, Jiang, Chen, Wang, Choi, and Lin}]{wildvision}
Yujie Lu, Dongfu Jiang, Wenhu Chen, William~Yang Wang, Yejin Choi, and Bill~Yuchen Lin. 2024.
\newblock \href {http://arxiv.org/abs/2406.11069} {Wildvision: Evaluating vision-language models in the wild with human preferences}.

\bibitem[{OpenAI(2025)}]{gpt5}
OpenAI. 2025.
\newblock \href {https://chat.openai.com/} {Gpt-5}.
\newblock Large language model.

\bibitem[{Radford et~al.(2021)Radford, Kim, Hallacy, Ramesh, Goh, Agarwal, Sastry, Askell, Mishkin, Clark et~al.}]{radford2021learning}
Alec Radford, Jong~Wook Kim, Chris Hallacy, Aditya Ramesh, Gabriel Goh, Sandhini Agarwal, Girish Sastry, Amanda Askell, Pamela Mishkin, Jack Clark, et~al. 2021.
\newblock Learning transferable visual models from natural language supervision.
\newblock In \emph{International conference on machine learning}, pages 8748--8763. PmLR.

\bibitem[{Rasley et~al.(2020)Rasley, Rajbhandari, Ruwase, and He}]{deepspeed}
Jeff Rasley, Samyam Rajbhandari, Olatunji Ruwase, and Yuxiong He. 2020.
\newblock Deepspeed: System optimizations enable training deep learning models with over 100 billion parameters.
\newblock In \emph{Proceedings of the 26th ACM SIGKDD International Conference on Knowledge Discovery \& Data Mining}, KDD '20, page 3505–3506, New York, NY, USA. Association for Computing Machinery.

\bibitem[{Sainz et~al.(2025)Sainz, Perez, Etxaniz, de~Landa, Aldabe, García-Ferrero, Zabala, Azurmendi, Rigau, Agirre, Artetxe, and Soroa}]{latxa}
Oscar Sainz, Naiara Perez, Julen Etxaniz, Joseba~Fernandez de~Landa, Itziar Aldabe, Iker García-Ferrero, Aimar Zabala, Ekhi Azurmendi, German Rigau, Eneko Agirre, Mikel Artetxe, and Aitor Soroa. 2025.
\newblock \href {http://arxiv.org/abs/2506.07597} {Instructing large language models for low-resource languages: A systematic study for basque}.

\bibitem[{Schwenk et~al.(2022)Schwenk, Khandelwal, Clark, Marino, and Mottaghi}]{aokvqa}
Dustin Schwenk, Apoorv Khandelwal, Christopher Clark, Kenneth Marino, and Roozbeh Mottaghi. 2022.
\newblock \href {https://doi.org/10.1007/978-3-031-20074-8_9} {A-okvqa: A benchmark for visual question answering using world knowledge}.
\newblock In \emph{Computer Vision – ECCV 2022: 17th European Conference, Tel Aviv, Israel, October 23–27, 2022, Proceedings, Part VIII}, page 146–162, Berlin, Heidelberg. Springer-Verlag.

\bibitem[{Sharma et~al.(2018)Sharma, Ding, Goodman, and Soricut}]{cc3m}
Piyush Sharma, Nan Ding, Sebastian Goodman, and Radu Soricut. 2018.
\newblock \href {https://doi.org/10.18653/v1/P18-1238} {Conceptual captions: A cleaned, hypernymed, image alt-text dataset for automatic image captioning}.
\newblock In \emph{Proceedings of the 56th Annual Meeting of the Association for Computational Linguistics (Volume 1: Long Papers)}, pages 2556--2565, Melbourne, Australia. Association for Computational Linguistics.

\bibitem[{Tong et~al.(2024)Tong, Brown, Wu, Woo, Middepogu, Akula, Yang, Yang, Iyer, Pan, Wang, Fergus, LeCun, and Xie}]{cambrian}
Shengbang Tong, Ellis Brown, Penghao Wu, Sanghyun Woo, Manoj Middepogu, Sai~Charitha Akula, Jihan Yang, Shusheng Yang, Adithya Iyer, Xichen Pan, Austin Wang, Rob Fergus, Yann LeCun, and Saining Xie. 2024.
\newblock \href {https://proceedings.neurips.cc/paper_files/paper/2024/file/9ee3a664ccfeabc0da16ac6f1f1cfe59-Paper-Conference.pdf} {Cambrian-1: A fully open, vision-centric exploration of multimodal llms}.
\newblock In \emph{Advances in Neural Information Processing Systems}, volume~37, pages 87310--87356. Curran Associates, Inc.

\bibitem[{Turisini et~al.(2023)Turisini, Amati, and Cestari}]{leonardo}
Matteo Turisini, Giorgio Amati, and Mirko Cestari. 2023.
\newblock \href {http://arxiv.org/abs/2307.16885} {Leonardo: A pan-european pre-exascale supercomputer for hpc and ai applications}.

\bibitem[{Yang et~al.(2025)Yang, Li, Yang, Zhang, Hui, Zheng, Yu, Gao, Huang, Lv, Zheng, Liu, Zhou, Huang, Hu, Ge, Wei, Lin, Tang, Yang, Tu, Zhang, Yang, Yang, Zhou, Zhou, Lin, Dang, Bao, Yang, Yu, Deng, Li, Xue, Li, Zhang, Wang, Zhu, Men, Gao, Liu, Luo, Li, Tang, Yin, Ren, Wang, Zhang, Ren, Fan, Su, Zhang, Zhang, Wan, Liu, Wang, Cui, Zhang, Zhou, and Qiu}]{qwen}
An~Yang, Anfeng Li, Baosong Yang, Beichen Zhang, Binyuan Hui, Bo~Zheng, Bowen Yu, Chang Gao, Chengen Huang, Chenxu Lv, Chujie Zheng, Dayiheng Liu, Fan Zhou, Fei Huang, Feng Hu, Hao Ge, Haoran Wei, Huan Lin, Jialong Tang, Jian Yang, Jianhong Tu, Jianwei Zhang, Jianxin Yang, Jiaxi Yang, Jing Zhou, Jingren Zhou, Junyang Lin, Kai Dang, Keqin Bao, Kexin Yang, Le~Yu, Lianghao Deng, Mei Li, Mingfeng Xue, Mingze Li, Pei Zhang, Peng Wang, Qin Zhu, Rui Men, Ruize Gao, Shixuan Liu, Shuang Luo, Tianhao Li, Tianyi Tang, Wenbiao Yin, Xingzhang Ren, Xinyu Wang, Xinyu Zhang, Xuancheng Ren, Yang Fan, Yang Su, Yichang Zhang, Yinger Zhang, Yu~Wan, Yuqiong Liu, Zekun Wang, Zeyu Cui, Zhenru Zhang, Zhipeng Zhou, and Zihan Qiu. 2025.
\newblock \href {http://arxiv.org/abs/2505.09388} {Qwen3 technical report}.

\bibitem[{Yue et~al.(2025)Yue, Song, Asai, Kim, de~Dieu~Nyandwi, Khanuja, Kantharuban, Sutawika, Ramamoorthy, and Neubig}]{pangea}
Xiang Yue, Yueqi Song, Akari Asai, Seungone Kim, Jean de~Dieu~Nyandwi, Simran Khanuja, Anjali Kantharuban, Lintang Sutawika, Sathyanarayanan Ramamoorthy, and Graham Neubig. 2025.
\newblock \href {https://openreview.net/forum?id=a3g2l4yEys} {Pangea: A fully open multilingual multimodal {LLM} for 39 languages}.
\newblock In \emph{The Thirteenth International Conference on Learning Representations}.

\bibitem[{Zhai et~al.(2023)Zhai, Mustafa, Kolesnikov, and Beyer}]{zhai2023sigmoid}
Xiaohua Zhai, Basil Mustafa, Alexander Kolesnikov, and Lucas Beyer. 2023.
\newblock Sigmoid loss for language image pre-training.
\newblock In \emph{Proceedings of the IEEE/CVF international conference on computer vision}, pages 11975--11986.

\bibitem[{Zhang et~al.(2024)Zhang, Yu, Dong, Li, Su, Chu, and Yu}]{MLLMReview}
Duzhen Zhang, Yahan Yu, Jiahua Dong, Chenxing Li, Dan Su, Chenhui Chu, and Dong Yu. 2024.
\newblock \href {https://doi.org/10.18653/v1/2024.findings-acl.738} {{MM}-{LLM}s: Recent advances in {M}ulti{M}odal large language models}.
\newblock In \emph{Findings of the Association for Computational Linguistics: ACL 2024}, pages 12401--12430, Bangkok, Thailand. Association for Computational Linguistics.

\bibitem[{Zhang et~al.(2025)Zhang, Li, Zhang, Pu, Cahyono, Hu, Liu, Zhang, Yang, Li, and Liu}]{lmmseval}
Kaichen Zhang, Bo~Li, Peiyuan Zhang, Fanyi Pu, Joshua~Adrian Cahyono, Kairui Hu, Shuai Liu, Yuanhan Zhang, Jingkang Yang, Chunyuan Li, and Ziwei Liu. 2025.
\newblock \href {https://doi.org/10.18653/v1/2025.findings-naacl.51} {{LMM}s-eval: Reality check on the evaluation of large multimodal models}.
\newblock In \emph{Findings of the Association for Computational Linguistics: NAACL 2025}, pages 881--916, Albuquerque, New Mexico. Association for Computational Linguistics.

\end{thebibliography}

\appendix
\section{Multimodal Large Language Model Training}\label{sec:Appendix2}

The two-stage training procedure requires different training configurations and infrastructure for each of the stages. For the Visual-Language Alignment stage, we trained the two configurations using a single node of 4 A100-sxm4 GPUs of the Hyperion cluster from the Donostia International Physics Center. Each training took a total of 80 GPU hours. This configuration led us to use a global batch size of 32 samples. We used the AdamW optimizer \cite{adamw} together with a cosine learning rate scheduler starting on an initial learning rate of $1e^{-3}$ and warm-up ratio of $0.03$. As for memory optimizations, we used gradient checkpointing and DeepSpeed ZeRO-3 \cite{deepspeed}.  These hyperparameters were based on \cite{pangea}, except that we opted for a context length of 8192 tokens, which results in a more VRAM-intensive training than the reference work.

For the Multimodal Instruction Tuning stage, training was conducted on the Leonardo supercomputer at CINECA’s HPC infrastructure \cite{leonardo} cluster. We utilize 8 nodes, providing a total of 32 A100 GPUs per training run. Following the Visual-Language Alignment phase, we maintained the context length of 8,192 tokens and employed gradient checkpointing for memory optimization. The training setup incorporated mixed-precision computation and DeepSpeed ZeRO-3 \cite{deepspeed} for fully sharded data parallelism across all GPUs. Given the batch size of 8 per node and gradient accumulation steps of 2, we achieve a global batch size of 128 samples. Following the Pangea methodology \cite{pangea}, we employed the AdamW optimizer with a cosine learning rate scheduler initialized at \num{2e-5}, weight decay of 0.01, and a warm-up ratio of 0.03. Each model trains for approximately 3-4 hours, depending on the specific dataset configuration.


\section{Translation of the Datasets}\label{sec:Appendix1}
Two translation methods have been used during this work. A sentence-level translation model for the CC3M dataset and an LLM-based translation procedure for the rest of the datasets and benchmarks. For the former, we have used the mt-hitz-en-eu model with the standard hyperparameters. Namely, no repetition penalty, temperature of 1.0, and top\_p of 1.0. As for the latter datasets, we adopted a similar instruction translation procedure used for text-only datasets in the Latxa work, using the Latxa-Llama-3.1-70B-Instruct LLM \cite{latxa}. The model was prompted to translate English question-answer pairs into Basque using a 2-shot configuration. The detailed prompts for each dataset are shown in Figures \ref{fig:translation-procedure1}, \ref{fig:translation-procedure2}, \ref{fig:translation-procedure3} and \ref{fig:translation-procedure4}.  

For inference with the Latxa model, we have followed the recommended hyperparameter settings. That is, a temperature of 0.7, no repetition penalty, and top\_p of 0.15. As for computational resources, we have used a single node of the Leonardo supercomputer at CINECA's HPC infrastructure \cite{leonardo}. The node is equipped with four NVIDIA Ampere A100 SXM4 GPUs (64GB HBM2e) interconnected via NVLink 3.0 and a 32-core Intel Xeon Platinum 8358 CPU (Ice Lake) with 512GB DDR4-3200 RAM. This node configuration allowed for a batch size of 64 question-answer pairs with a context length of 2048 tokens, which we found to be enough for all the samples.

\begin{figure*}[htbp]

\centering
\begin{tcolorbox}[
    colback=customblue!10,
    colframe=customblue,
    boxrule=1pt,
    arc=3pt,
    left=6pt, right=6pt, top=6pt, bottom=6pt,
    title=Translation Procedure for \textbf{A-OKVQA\textsubscript{Eus}}: 2-Shot Prompting for English-Basque,
    width=0.95\textwidth]

\begin{tcolorbox}[
    colback=customblue!5,
    colframe=customblue!70,
    boxrule=1pt,
    arc=3pt,
    left=6pt, right=6pt, top=6pt, bottom=6pt,
    title=System Message,
    fonttitle=\bfseries]
\footnotesize
\textbf{Role:} System\\
\textbf{Content:} You are a helpful AI assistant that specializes in English to Basque translations.\textbackslash n Your task is to translate instruction datasets from English to Basque.\textbackslash n\textbackslash n Here are some important guidelines:\textbackslash n 1. Maintain the original meaning and intent of the instructions\textbackslash n 2. Use standard Basque language (batua)\textbackslash n 3. Keep technical terms that don't have widely accepted Basque translations\textbackslash n 4. Preserve any code snippets, variables, or special characters exactly as they appear\textbackslash n\textbackslash n Please provide accurate Basque translations for all text fields.
\end{tcolorbox}

\vspace{0.3cm}

\begin{tcolorbox}[
    colback=customblue!8,
    colframe=customblue!80,
    boxrule=1pt,
    arc=3pt,
    left=6pt, right=6pt, top=6pt, bottom=6pt,
    title=Shot 1,
    fonttitle=\bfseries]

\footnotesize
\textbf{Role:} User\\
\textbf{Content:} Translate the following question and answer to Basque \textbackslash n \textbf{Question}: What is the man by the bags awaiting? \textbackslash n \textbf{Answer}: 1.skateboarder, 2.train, 3.delivery, 4.cab

\vspace{0.3cm}

\textbf{Role:} Assistant\\
\textbf{Content:} \textbf{Galdera:} Zer ari da poltsekin dagoen gizona itxaroten?\textbf{Erantzuna:}  1.skaterra, 2.trena, 3.Deliveria, 4.Taxia
\end{tcolorbox}

\vspace{0.3cm}

\begin{tcolorbox}[
    colback=customblue!8,
    colframe=customblue!80,
    boxrule=1pt,
    arc=3pt,
    left=6pt, right=6pt, top=6pt, bottom=6pt,
    title=Shot 2,
    fonttitle=\bfseries]

\footnotesize
\textbf{Role:} User\\
\textbf{Content:} \textbf{Question:} Where does this man eat pizza? \textbf{Answer:} 1.office, 2.cafe, 3.motel, 4.outside
\vspace{0.3cm}

\textbf{Role:} Assistant\\
\textbf{Content:} \textbf{Galdera:} Non ari da gizon hau pizza jaten? \textbf{Erantzuna:} 1.Ofizinan, 2.Kafetegian, 3.Motelean, 4.Kanpoan
\end{tcolorbox}

\vspace{0.3cm}

\begin{tcolorbox}[
    colback=customblue!12,
    colframe=customblue!90,
    boxrule=1pt,
    arc=3pt,
    left=6pt, right=6pt, top=6pt, bottom=6pt,
    title=New Query,
    fonttitle=\bfseries]

\footnotesize
\textbf{Role:} User\\
\textbf{Content:} Translate the following question and answers to Basque \textbackslash n\textbackslash n\textbackslash n \textbf{Question}: \{example['question']\} \textbf{Answer}: \{example['answer']\}

\vspace{0.2cm}

\textbf{Role:} Assistant\\
\textbf{Content:} \textbf{Galdera:} ... \textbackslash n \textbf{Erantzuna:} ...

\end{tcolorbox}

\end{tcolorbox}
\caption{Two-shot prompting procedure for English to Basque translation of the A-OKVQA\textsubscript{Eus} benchmark. {example['question']} and {example['answer']} correspond to the question and answer to be translated in the example. }
\label{fig:translation-procedure1}
\end{figure*}
\begin{figure*}[htbp]

\centering
\begin{tcolorbox}[
    colback=customblue!10,
    colframe=customblue,
    boxrule=1pt,
    arc=3pt,
    left=6pt, right=6pt, top=6pt, bottom=6pt,
    title=Translation Procedure for \textbf{PixMo-CapQA\textsubscript{Eus}}: 2-Shot Prompting for English-Basque,
    width=0.95\textwidth]

\begin{tcolorbox}[
    colback=customblue!5,
    colframe=customblue!70,
    boxrule=1pt,
    arc=3pt,
    left=6pt, right=6pt, top=6pt, bottom=6pt,
    title=System Message,
    fonttitle=\bfseries]
\footnotesize
\textbf{Role:} System\\
\textbf{Content:} You are a helpful AI assistant that specializes in English to Basque translations.\textbackslash n Your task is to translate instruction datasets from English to Basque.\textbackslash n\textbackslash n Here are some important guidelines:\textbackslash n 1. Maintain the original meaning and intent of the instructions\textbackslash n 2. Use standard Basque language (batua)\textbackslash n 3. Keep technical terms that don't have widely accepted Basque translations\textbackslash n 4. Preserve any code snippets, variables, or special characters exactly as they appear\textbackslash n\textbackslash n Please provide accurate Basque translations for all text fields.
\end{tcolorbox}

\vspace{0.3cm}

\begin{tcolorbox}[
    colback=customblue!8,
    colframe=customblue!80,
    boxrule=1pt,
    arc=3pt,
    left=6pt, right=6pt, top=6pt, bottom=6pt,
    title=Shot 1,
    fonttitle=\bfseries]

\footnotesize
\textbf{Role:} User\\
\textbf{Content:} Translate the following question and answer to Basque \textbackslash n \textbf{Question}: Is this room decorated for the 1970s?

\vspace{0.3cm}

\textbf{Role:} Assistant\\
\textbf{Content:} \textbf{Galdera:} Gela hau 1970ean bezala dago diseinatuta?
\end{tcolorbox}

\vspace{0.3cm}

\begin{tcolorbox}[
    colback=customblue!8,
    colframe=customblue!80,
    boxrule=1pt,
    arc=3pt,
    left=6pt, right=6pt, top=6pt, bottom=6pt,
    title=Shot 2,
    fonttitle=\bfseries]

\footnotesize
\textbf{Role:} User\\
\textbf{Content:} Translate the following question and answers to Basque \textbackslash n \textbf{Question:} Is there a mirror in the room?
\vspace{0.3cm}

\textbf{Role:} Assistant\\
\textbf{Content:} \textbf{Galdera:} Ba al dago ispilurik gelan?
\end{tcolorbox}

\vspace{0.3cm}

\begin{tcolorbox}[
    colback=customblue!12,
    colframe=customblue!90,
    boxrule=1pt,
    arc=3pt,
    left=6pt, right=6pt, top=6pt, bottom=6pt,
    title=New Query,
    fonttitle=\bfseries]

\footnotesize
\textbf{Role:} User\\
\textbf{Content:} Translate the following question and answers to Basque \textbackslash n\textbackslash n\textbackslash n \textbf{Question}: \{example['question']\} \textbf{Answer}: \{example['answer']\}

\vspace{0.2cm}

\textbf{Role:} Assistant\\
\textbf{Content:} \textbf{Galdera:} ... \textbackslash n \textbf{Erantzuna:} ...

\end{tcolorbox}

\end{tcolorbox}
\caption{Two-shot prompting procedure for English to Basque translation of the PixMo-CapQA\textsubscript{Eus} benchmark. {example['question']} and {example['answer']} correspond to the question and answer to be translated in the example. }
\label{fig:translation-procedure2}
\end{figure*}
\begin{figure*}[htbp]

\centering
\begin{tcolorbox}[
    colback=customblue!10,
    colframe=customblue,
    boxrule=1pt,
    arc=3pt,
    left=6pt, right=6pt, top=6pt, bottom=6pt,
    title=Translation Procedure for \textbf{VQAv2\textsubscript{Eus}}: 2-Shot Prompting for English-Basque,
    width=0.95\textwidth]

\begin{tcolorbox}[
    colback=customblue!5,
    colframe=customblue!70,
    boxrule=1pt,
    arc=3pt,
    left=6pt, right=6pt, top=6pt, bottom=6pt,
    title=System Message,
    fonttitle=\bfseries]
\footnotesize
\textbf{Role:} System\\
\textbf{Content:} You are a helpful AI assistant that specializes in English to Basque translations.\textbackslash n Your task is to translate instruction datasets from English to Basque.\textbackslash n\textbackslash n Here are some important guidelines:\textbackslash n 1. Maintain the original meaning and intent of the instructions\textbackslash n 2. Use standard Basque language (batua)\textbackslash n 3. Keep technical terms that don't have widely accepted Basque translations\textbackslash n 4. Preserve any code snippets, variables, or special characters exactly as they appear\textbackslash n\textbackslash n Please provide accurate Basque translations for all text fields.
\end{tcolorbox}

\vspace{0.3cm}

\begin{tcolorbox}[
    colback=customblue!8,
    colframe=customblue!80,
    boxrule=1pt,
    arc=3pt,
    left=6pt, right=6pt, top=6pt, bottom=6pt,
    title=Shot 1,
    fonttitle=\bfseries]

\footnotesize
\textbf{Role:} User\\
\textbf{Content:} Translate the following question and answer to Basque \textbackslash n \textbf{Question}: Explain the visual content of the image in great detail.\textbackslash n \textbf{Answer}: In this outdoor wedding photo, a bride wearing a beautiful white dress is posing on a park bench surrounded by a group of five men dressed in suits. The groomsmen are wearing pink ties, adding a pop of color to the scene. The group is standing around the bride, smiling and enjoying the special occasion as they pose for the picture.\textbackslash n \textbackslash n. In the background, there are potted plants which further enhance the ambiance of the park setting. The bench, on which the bride is seated, is located close to the center of the image.

\vspace{0.3cm}

\textbf{Role:} Assistant\\
\textbf{Content:} \textbf{Galdera:} Azaldu irudia zehaztasun askorekin. \textbf{Erantzuna:}  Ezkontzako argazki honetan, soineko zuri eder bat daraman emaztegaia parkeko banku batean dago, trajez jantzitako bost gizonez inguratuta. Neskameek gorbata arrosak daramatzate, eszenari kolorezko pop bat erantsiz. Taldea emaztegaiaren inguruan dago, irribarretsu, eta aukera bereziaz gozatzen dute koadrorako. Jarlekua, non emaztegaia eserita dagoen, irudiaren erditik hurbil dago.
\end{tcolorbox}

\vspace{0.3cm}

\begin{tcolorbox}[
    colback=customblue!8,
    colframe=customblue!80,
    boxrule=1pt,
    arc=3pt,
    left=6pt, right=6pt, top=6pt, bottom=6pt,
    title=Shot 2,
    fonttitle=\bfseries]

\footnotesize
\textbf{Role:} User\\
\textbf{Content:} Translate the following question and answers to Basque \textbackslash n \textbf{Question:} What website copyrighted the picture? \textbf{Answer:} 1.foodiebakercom, 2.foodiebakercom, 3.foodiebaker, 4.foodiebakercom, 5.foodiebakercom, 6.http://foodiebakercom, 7.foodiebakercom, 8.foodiebakercom, 9.foodiebakercom, 10.foodiebaker
\vspace{0.3cm}

\textbf{Role:} Assistant\\
\textbf{Content:} \textbf{Galdera:} Zein webgunek du irudiaren copyrigheta? \textbf{Erantzuna:} 1.foodiebakercom, 2.foodiebakercom, 3.foodiebaker, 4.foodiebakercom, 5.foodiebakercom, 6.http://foodiebakercom, 7.foodiebakercom, 8.foodiebakercom, 9.foodiebakercom, 10.foodiebaker
\end{tcolorbox}

\vspace{0.3cm}

\begin{tcolorbox}[
    colback=customblue!12,
    colframe=customblue!90,
    boxrule=1pt,
    arc=3pt,
    left=6pt, right=6pt, top=6pt, bottom=6pt,
    title=New Query,
    fonttitle=\bfseries]

\footnotesize
\textbf{Role:} User\\
\textbf{Content:} Translate the following question and answers to Basque \textbackslash n\textbackslash n\textbackslash n \textbf{Question}: \{example['question']\} \textbf{Answer}: \{example['answer']\}

\vspace{0.2cm}

\textbf{Role:} Assistant\\
\textbf{Content:} \textbf{Galdera:} ... \textbackslash n \textbf{Erantzuna:} ...

\end{tcolorbox}

\end{tcolorbox}
\caption{Two-shot prompting procedure for English to Basque translation of the VQAv2\textsubscript{Eus} benchmark. {example['question']} and {example['answer']} correspond to the question and answer to be translated in the example.}
\label{fig:translation-procedure3}
\end{figure*}
\begin{figure*}[htbp]

\centering
\begin{tcolorbox}[
    colback=customblue!10,
    colframe=customblue,
    boxrule=1pt,
    arc=3pt,
    left=6pt, right=6pt, top=6pt, bottom=6pt,
    title=Translation Procedure for \textbf{Wildvision\textsubscript{Eus}}: 2-Shot Prompting for English-Basque,
    width=0.95\textwidth]

\begin{tcolorbox}[
    colback=customblue!5,
    colframe=customblue!70,
    boxrule=1pt,
    arc=3pt,
    left=6pt, right=6pt, top=6pt, bottom=6pt,
    title=System Message,
    fonttitle=\bfseries]
\footnotesize
\textbf{Role:} System\\
\textbf{Content:} Cutting Knowledge Date: December 2023
Today Date: [2025-01-01]

You are a helpful assistant
\end{tcolorbox}

\vspace{0.3cm}

\begin{tcolorbox}[
    colback=customblue!8,
    colframe=customblue!80,
    boxrule=1pt,
    arc=3pt,
    left=6pt, right=6pt, top=6pt, bottom=6pt,
    title=Shot 1,
    fonttitle=\bfseries]
\footnotesize
\textbf{Role:} User\\
\textbf{Content:} Honako testu hau gaztelaniaz idatzita dago eta euskerara itzuli behar da. Itzulpena zehatza eta naturala izan behar da. Jarri ezazu bakarrik testua, ez egin erreferentziarik hau itzulpen bat denik.

\textbf{Gaztelaniazko testua:} example['question']

\textbf{Euskerazko itzulpena:}

\vspace{0.3cm}

\textbf{Role:} Assistant\\
example['answer']
\end{tcolorbox}
\end{tcolorbox}
\caption{Prompt for the Spanish to Basque translation of the Wildvision\textsubscript{Eus} benchmark. {example['question']} and {example['answer']} correspond to the question and answer to be translated in the example. We used a 0-shot strategy for open-ended settings. }
\label{fig:translation-procedure4}
\end{figure*}

\end{document}